\documentclass[twocolumn]{article}
\usepackage{cite}
\usepackage{amsmath,amssymb,amsfonts}
\usepackage{algorithm}
\usepackage{multirow}
\usepackage{authblk}
\usepackage{algpseudocode}
\usepackage{graphicx}
\usepackage{textcomp}
\def\BibTeX{{\rm B\kern-.05em{\sc i\kern-.025em b}\kern-.08em
    T\kern-.1667em\lower.7ex\hbox{E}\kern-.125emX}}

\usepackage[utf8]{inputenc}
\usepackage[T1]{fontenc}
\usepackage{csquotes}

\begin{document}
\title{Inference of Causal Networks using a Topological Threshold}

\author[1]{Filipe Barroso\thanks{filipe.barroso@ua.pt}}
\author[2,3]{Diogo Gomes}
\author[1]{Gareth J. Baxter}

\affil[1]{Department of Physics and i3N, University of Aveiro, 3810-193 Aveiro, Portugal}
\affil[2]{Instituto de Telecomunicações, Campus Universitário de Santiago, Aveiro, 3810-193, Portugal}
\affil[3]{Department of Electronics, Telecommunications and Informatics, University of Aveiro, Campus Universitário de Santiago, Aveiro, 3810-193, Portugal}

\maketitle


\begin{abstract}
We propose a constraint-based algorithm, which automatically determines causal relevance thresholds, to infer causal networks from data. We call these topological thresholds. We present two methods for determining the threshold: the first seeks a set of edges that leaves no disconnected nodes in the network; the second seeks a causal large connected component in the data.

We tested these methods both for discrete synthetic and real data, and compared the results with those obtained for the PC algorithm, which we took as the benchmark. We show that this novel algorithm is generally faster and more accurate than the PC algorithm.

The algorithm for determining the thresholds requires choosing a measure of causality. We tested our methods for Fisher Correlations, commonly used in PC algorithm (for instance in \cite{kalisch2005}), and further proposed a discrete and asymmetric measure of causality, that we called Net Influence, which provided very good results when inferring causal networks from discrete data. This metric allows for inferring directionality of the edges in the process of applying the thresholds, speeding up the inference of causal DAGs.
\end{abstract}

\maketitle

\section{Introduction}
Understanding causality in complex systems allows for the identification of root causes, prediction of the effects of interventions, or the establishment of causal laws to describe a system.
These considerations are relevant to areas such
industry \cite{vukovic2022causal}, telecommunications \cite{zhang2020influence},
medicine \cite{ramazzotti2018modeling},
environmental studies \cite{perdicoulis2006causal},
earth system sciences \cite{runge2019inferring}
and meteorology \cite{cano2004applications}.

A convenient way to encode causal relations between elements in a complex system is a causal network, in which edges represent some causal relationship between two events, and the absence of an edge between variables indicates causal independence, or d-separation \cite{pearl1985bayesian,Pearl2009,geiger2013dseparation}. For the purpose of this work, multiple edges to a single node should be understood as disjunct causal relations. However, more complex descriptions, such as causal hypergraphs, can be used to describe conjunction of causality \cite{Ausiello2017}.

Time ordering of events imposes a directionality on the edges, and the impossibility of retrocausality implies causal networks are acyclic. These two conditions can be summarized mathematically by describing networks as Directed Acyclic Graphs (DAGs). Bayesian networks represent the network factorization of the joint probability function of the system and are generally good approximations for causation.

Bayesian networks might be established through experimentation. However, due to practical, financial, or ethical constraints, we may be unable to freely experiment with the system. Thus, causal networks often need to be inferred from collected data alone.
Furthermore, a system may contain many variables, implying a very large set of potential influences to check. Manual construction of a causal network is often also impractical, requiring automated methods instead.

In man-made systems such as industrial production \cite{vukovic2022causal} or telecommunications \cite{zhang2020influence}, we may have access to large datasets of sample system states, i.e. the states of all or some variables at a given time. This opens up the possibility for automated inference of the causal network based on data. Considering the possible practical and time constraints, inferring a network of causal relations solely from collected data is a problem of special importance.

The relationship between graphs and probabilistic dependencies was established in the 1980s \cite{Pearl2009}. Given a correct ordering of the nodes, the underlying causal structure can be correctly recovered, but it requires testing statistical independence between any two nodes given all the possible separating sets \cite{geiger2013dseparation,Verma_Pearl}. Such a task is computationally infeasible for large networks. Furthermore, it might be impossible to calculate conditional measures with sufficient precision given finite datasets, especially when the number of dependencies is large. Therefore, practical algorithms try to reach a compromise between correctness and computability \cite{zheng2018dags,Le2019,qi2019learning,sun2015causal,fang2023efficient}.

The inference of a Bayesian network can be divided into two main tasks: identifying the correct topology of the network, and determining the influence strengths. Here we focus on the first task. Once the structure is decided, the second task becomes one of calculating conditional probabilities from data.

There are two main approaches for learning networks: score-based methods and constraint-based methods. Score-based methods assign a score to each Bayesian structure according to a scoring function and optimizes the score \cite{Liu2012}. A example of a method in this category is the Belief Learning Network \cite{Cooper_Herskovits}. A derived method for continuous variables that avoids the large combinatorial problem was recently proposed by \cite{zheng2018dags}, where the authors propose a smooth characterization of acyclicity and use continuous optimization.

On the other hand, constraint-based methods construct the graph according to some measure of conditional dependence. Some such methods start with complete graphs and remove edges through the evaluation of conditional independence (CI). The most well-known such method, the PC algorithm, deletes edges recursively \cite{Spirtes2000}. These methods, for large number of variables, are computationally impractical, so they can only be applied approximately. Several improvements of the PC algorithm have been made, such as parallelization \cite{Le2019} or strategical edge removal order \cite{qi2019learning}.

Other methods build up the graph step by step, based on local evaluations of CI relative to a given variable, followed by refinement of the graph. Examples include the grow-shrink algorithm \cite{margaritis1999bayesian} based on $\chi^2$ conditional tests, further developed in \cite{bromberg2009efficient},
and the entropy based method developed in \cite{sun2015causal}.
Reference \cite{cheng2002learning} computes pairwise CI in bulk, then uses a threshold to grow an approximate network based on pairwise independence measures, and then trims spurious correlations by comparing measures of CI against the threshold.

Yet other methods use a hybrid of constraint- and score-based methods \cite{tsamardinos2006max}, and more recent works also apply machine learning methods \cite{fang2023efficient}.

With finite datasets, it is never possible to establish probabilities exactly. Furthermore, it is counterproductive to represent every dependency, no matter how small, in the causal network. It is therefore necessary to establish a significance threshold for conditional dependence. Naturally, the correct threshold depends on the system being studied, on the variable types and on the structure of the data. As noted by\cite{natori2017}, previous works have tended to gloss over this point, choosing the threshold in an {\em ad-hoc} way \cite{cheng2002learning, claassen2012bayesian}.

In systems in which variables have a finite number of discrete states, it may be advantageous to conduct the analysis at the resolution of the variable state, not the aggregate variable. If one later wishes to use the obtained network to analyse and predict the evolution of the system state, or infer root causes of certain states, it makes sense to consider the state as the fundamental unit. The NI measure we define is designed specifically to address this case.

In this paper, we demonstrate that constraint-based algorithms can be freed from the fine-tuning of thresholds. We show that computation of pairwise measures combined with a topological criterion that seeks to connect nodes to the largest connected component (LCC), suffices to determine an appropriate threshold. We present two options: strictly connecting all nodes in one LCC, or finding the largest component which includes most nodes while avoiding weak connections whose addition could introduce a large number of spurious correlations. This second approach works by finding the knee point in the curve of size of the largest network component vs ranked edges.

We show that these thresholds can be efficiently used as topological criteria for constraints, with their application presenting very good results. We show that for general networks, removing unconditional independence and CI to a single node outperforms the PC algorithm, which we take as the benchmark, in time and quality of inference.

In principle, our algorithm can be adapted to any measure of CI. In fact, there is no uniformly valid CI test \cite{shah2020hardness,watson2021testing}. However, in this work we further introduce a new probabilistic, and asymmetric, measure of CI, that we dub "Net influence" (NI). It is a modified form of certainty factor \cite{shortliffe1975model}. We show that, at least in conjunction with our algorithm, NI can be used to infer the directionality of causal influences for discrete data.

In the next section we describe our use-case and define Net Influence. In section \ref{Algorithm} we present our innovative inference algorithm and do a complexity analysis.
In section \ref{Real networks} we demonstrate our algorithm with some example real-world networks. In section \ref{Synthetic networks} we generalize the study and show that our algorithm achieves good results for a large set of synthetically generated networks and data, with randomly assigned influence. We conclude the paper in section \ref{Conclusion} with a discussion on the impact and usefulness of the proposed algorithm. 

\section{Inference of causal networks}

Consider a complex system defined by a set of variables. For each variable we have a set of dataset of measurements of their states. We wish to infer the structure of the causal network, $G(V,E)$, where nodes $V$ correspond to variables and (directed) edges $E$ represent causal influence. For clarity, we will refer to the source of an edge as the parent node, and the destination of the edge as the child node.
We assume that each line of data consists of a set of measurements of the states of the variables at a particular moment. Thus each line represents a realisation of the state of the system. We do not explicitly consider time series data, instead treating each line of data as an independent, Markovian, record. Knowledge of temporal ordering, which simplifies the problem of discovering the direction of the edges, is not assumed, but the algorithm can be easily adapted to include this additional information.

Heuristically, we expected that a significant part of the system we wish to infer is part of a single connected component. We defined significance thresholds based on this property. The first option assumes all variables are connected and we solely seek to find the connections between then, thus looking for maximum threshold that infers a single connected component. In certain cases, however, it may not lead to a sensible representation of the system. For instance, if there are nodes or small components which only very weakly influence or are influenced by other nodes in the system, their inclusion may force an artificially low threshold, creating too dense a network, which is computationally costly to analyse. Thus, we present a second option which uses the size of the largest network component as an indicator to determine the significance threshold for the inferred network, looking for the balance between size of the largest component and number of edges included.

Let us consider a causal node, $I\in V$, as a statistical variable. If $I$ is discrete, in a given measurement, it can assume one of several node states, $i\in I$. We consider the existence of a causal edge, $J\to I\in E$, when there is a direct and measurable influence of a state $j\in J$ on $i\in I$, given the state of other nodes which may also possibly influence $i$. For continuous variables, this corresponds to a correlation between the values of $J$ and $I$, and the argument still applies. Statistical independence implies the absence of a causal relation between two nodes.

In this work, we tested the inference method for two measures of influence: a continuous measure, Fisher transformed Pearson correlation, used by PC algorithm for instance in \cite{kalisch2005}, and a new, state-wise, measure that we named Net Influence, defined below.

We used the measures to infer both the skeleton of the network and its DAG. We note that, contrary to the PC algorithm, which first identifies the skeleton and only later decides the directions, we sought to infer directionality the moment we apply the threshold.

\subsection{Net Influence}
Let $I$ be a node whose parents are $J$ and the set of nodes $U$. Let $i$ be a state of $I$, $j$ a state of $J$, and $u$ a set consisting of one state from each node of $U$. We introduce a quantity used to measure the influence of $j\in J$ on $i\in I$ in the presence of the set of other influences $u\in U$ as
\begin{equation}\label{eq:A}
    \begin{split}
        W\left(i\mid j;u\right) & = P\left(i\mid j, u\right) - P\left(i\mid \bar{j}, u\right) \\
        & = \frac{P\left(i,j\mid u\right)-P\left(i\mid u\right)P\left(j\mid u\right)}{P\left(j\mid u\right)P\left(\bar{j}\mid u\right)}
    \end{split}
\end{equation}
in which $\bar{j}$ indicates the absence of state $j$. We call this measure Net Influence (NI).

\begin{figure}
    \centering
    \includegraphics[width=6cm]{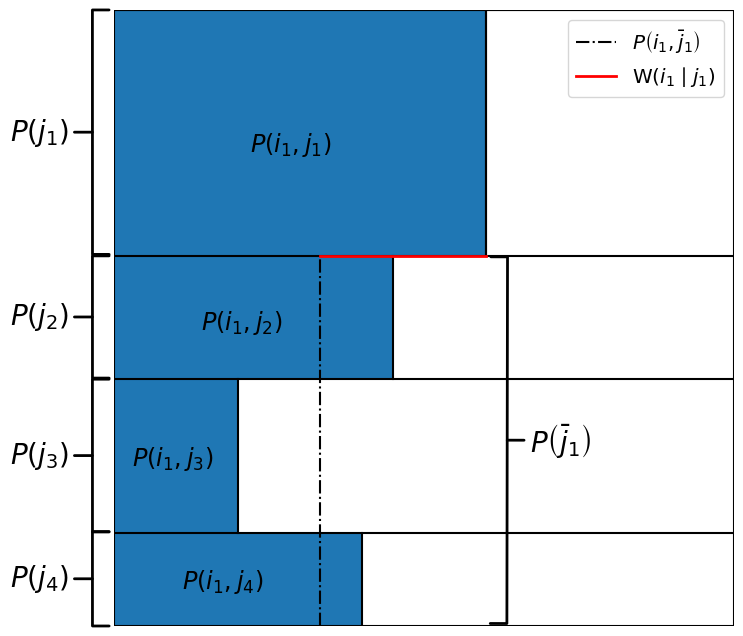}
    \caption{An illustration of the influence of state $j_1\in J$ on state $i_1 \in I$, as measured by Net Influence $W\left(i_1\mid j_1\right)$. Net Influence is the difference between the probability of state $i_1$ in the presence of $j_1$ from the probability of $i_1$ under any other state of $J$, and can be decomposed as a weighted mean of the differences of the $P\left(i_1\mid j_k\right)$.}
    \label{fig:ni}
\end{figure}

Net influence takes values in $[-1,1]$ and yakes a null value when $i$ and $j$ are independent given $u$. The extreme values are obtained when $i$ is fully determined by $j$ or by its inverse, respectively.

Since $P\left(i\mid u\right) = \sum_{j^\prime\in J} P\left(j^\prime\mid u\right) P\left(i\mid j^\prime, u\right)$, Equation \eqref{eq:A} can be written as
\begin{equation}\label{eq:B}
    W\left(i\mid j;u\right) = \sum_{j^\prime\in J}\frac{P\left(j^\prime\mid u\right)}{P\left(\bar{j}\mid u\right)}\bigl[P\left(i\mid j,u\right) - P\left(i\mid j^\prime,u\right) \bigr],
\end{equation}
with the sum running over all states of $J$. Defining $d_{i\mid u}\left(j,j^\prime\right) = P\left(i\mid j, u\right) - P\left(i\mid j^\prime, u\right)$ as the influence distance between states $j,j^\prime\in J$ on $i$ in the presence of $u$, allows us to rewrite Equation \eqref{eq:B} as
\begin{equation}
    \begin{split}        
    W\left(i\mid j;u\right) &= \frac{\sum_{j^\prime\in J}P\left(j^\prime\mid u\right)d_{i\mid u}\left(j,j^\prime\right)}{1-P\left(j\mid u\right)} \\
    &= \frac{\sum_{j^\prime\in J\setminus \{j\}}P\left(j^\prime\mid u\right)d_{i\mid u}\left(j,j^\prime\right)}{\sum_{j^\prime\in J\setminus \{j\}} P\left(j^\prime\mid u\right)}.
    \end{split}
\end{equation}

Thus, the net influence $W\left(i\mid j;u\right)$ can be interpreted as the weighted mean of the influence distances on state $i\in I$ from the node $J$, in the presence of $u$ (see Figure \ref{fig:ni}). In the simpler case where $J$ has only two states, $j,\bar{j}$, $W\left(i\mid j;u\right)=d_{i\mid u}\left(j,\bar{j}\right)$.

\section{Inference algorithm}\label{Algorithm}

Evaluating all combinations of statistical dependencies should, in theoretically perfect conditions, yield the correct skeleton \cite{geiger2013dseparation,Verma_Pearl}. In practice, computational feasibility demands inventive algorithms to reduce the number of evaluations of CI. Furthermore, statistical fluctuations make it impossible to perfectly distinguish real and false edges, therefore a significance threshold is needed.

The number of computations of CI can be reduced by evaluating them sequentially. In a given $n$th step, CI can be evaluated for all edges conditioned to $n$ of their common neighbours (whenever they exist), removing the edge when one of the conditioning set leaves the pair of variables below the threshold for statistical independence. The zeroth step evaluates the unconditioned independence of all pairs of nodes.

Our method for inferring the causal network behind data consists of two stages: an automatic determination of the threshold based on provided data, and the removal of statistically weak edges starting from the complete graph, that is, the imposition of constraints on the possible causal connections based on the determined threshold.

The threshold is determined using the statistical (unconditioned) dependencies between pairs of nodes, as these quantities can be viewed as the leading order of the CIs evaluations. Crucially, since the values of CI needed to determine the threshold are the ones used for pruning the graph in zeroth order, our method of determining the threshold avoids introducing a large number of additional computations.

We find that completing the second stage to the zeroth and first order CI evaluations yields very good results, saving computational time. However, in principle the method can easily be expanded to higher order.

\subsection{First stage}
A threshold is needed in order to keep only the relevant causal relations and cut out nonzero CIs arising solely from statistical fluctuations.
Previous works used a manually chosen threshold at the equivalent step \cite{cheng2002learning, claassen2012bayesian, natori2017}. However different systems with different levels of noise, types of variables and so on, will naturally require different thresholds.

We start this stage by obtaining an estimate of the influence, or statistical dependence, by considering only interactions between pairs of variables (i.e. not conditioned on other variable states). If a variable is discretised in states a measure of influence $\omega_{ij}$ can be computed between each pair of states of different nodes and the maximum value for a pair of nodes is considered its maximum potential influence, which we designate the weight $\Omega_{IJ}$ of the edge, serving as the CI measure. In principle a number of different CI measures could be used to estimate the influence between nodes. Here we focus on Net Influence, described above, and compare the results with the non-discrete Fisher correlation.

Only edges whose measure $\Omega_{IJ}$ is above the threshold should be included in the network.
 It is reasonable to assume that the majority (or even the totality) of variables form part of a single causal system. We present two methods to automatically choose the threshold based on topology of the leading order of CIs inferred from the provided data that connects all (or nearly all) of the variables. We call these topological thresholds.

\subsubsection{Connected method}
The first method to determine the threshold is called the Connected Method. In this method, the threshold is chosen such that all variables have at least one link to the largest (and only) connected component. This method reflects the cases when the user knows that all variables have a causal role in the system and wants to figure out their relations.

In order to apply this method, we find the maximum value of measure of influence for which removal of all edges below this threshold still leaves all nodes in the network connected.
The threshold is the solution to the following equations:
\begin{equation}
    \begin{array}{rcl}
        G_{\varepsilon} & = & \{(I,J) \mid \Omega_{IJ}>\varepsilon \} \\
        \varepsilon & = & \max({\varepsilon' \mid \mbox{LCC}(G_{\varepsilon'})  \ni I\ \forall\  I})
    \end{array}
\end{equation}
where $G_{\varepsilon}$ is the graph generated when using threshold value $\varepsilon$, and LCC indicates the largest connected component of the graph. This search can be quickly concluded through algorithms such as Binary Search. When inferring a DAG, $(I,J)$ denotes a directed edge, whereas for inferring a skeleton, $(I,J)$ denotes an undirected edge.

The application of this threshold in the second stage oftentimes yields a graph that includes the strongest causal influences in the system, being more accurate for tree-like causal systems. Sometimes, however, some variables might be only weakly connected with the main component of the network. Requiring the inclusion of these nodes forces the threshold to be much lower, resulting in large numbers of excess edges that are hard and time-consuming to remove in higher steps.

\begin{figure}[h!]
    \centering
    \includegraphics[width=\columnwidth]{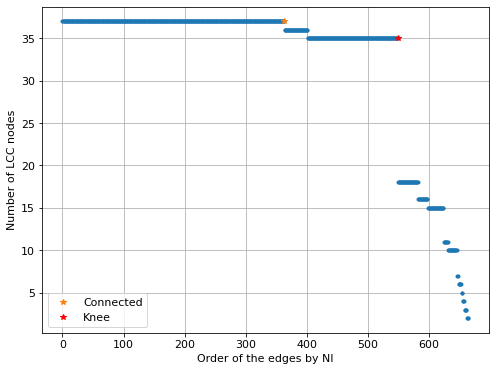}
    \caption{Number of nodes in the largest component of the network, as a function of number of edges removed, for the ALARM network (analysed below). The Connected method finds the point before a node drops from the LCC (orange marker). The Knee method finds the point of greatest curvature (red marker).}
    \label{fig:alarmgcc}
\end{figure}

\subsubsection{Knee method}\label{knee}
The problem that sometimes plagues the Connected method can be solved by looking instead at the number of nodes in the largest connected component (LCC) as the weaker edges are removed and finding the point of maximum curvature, or `knee’. Thus, the Knee method might drop some nodes from the inferred network, but we are able to present a cleaner causal network representing the relations in the LCC. Note that, in many instances, we expect the knee point to coincide with the point the LCC ceases to contain all nodes, reproducing the Connected method. In order to determine the knee point we used the algorithm developed in \cite{Mario2018}. The disadvantage of using the Knee method is that it may remove edges that should be included, that is, it creates more false negatives. As it does not use Binary Search, oftentimes this method might be slower than the Connected method. See Figure \ref{fig:alarmgcc} for a graphical comparison of Knee and Connected methods.



\subsection{Second stage}
Spurious correlations are statistically independent when conditioned on a certain, possibly empty, separating set. Removing these correlations is a matter of computing CI and comparing them with the relevance threshold. This is called constraining the graph. Crucially, the thresholds identified solely through the computation of unconditional independence measures serve as the relevance criteria for any CI test.

Although ideally we would try to identify all spurious correlations by computing conditioned independence for any separating set, for large networks this is computationally infeasible, even in the case when the zeroth step performs well in identifying the causal links. Instead, we apply an iterative method of, in any $n$th step, evaluating CIs conditioned to the existing $n$ parent nodes and edges evaluated as conditionally independent. This method is dominated by the zeroth step, whose computations were already obtained when determining the threshold. The first step, for instance, identifies edges whose nodes have at least one common parent and evaluates independence conditioned to each of those parents.

This approach greatly reduces the number of conditioned measures which need to be computed. Furthermore, considering that our calculations for NI are made state-by-state, the probabilities of combinations of the states of more than three variables may require much more data to identify accurately.

Note that edges wrongly removed - due to a too high threshold - are not recovered in following steps.

\begin{algorithm}
\caption{Algorithm}\label{alg:cap}
\begin{algorithmic}
\State \# \textit{Find threshold}
\For{variable $I,J$}
    \For{states $i\in I, j\in J$}
        \State $\omega_{ij} \gets |W\left(i\mid j\right)|$;
    \EndFor
    \State $\Omega_{IJ} \gets \max{\left(\omega_{ij}\right)}$;
\EndFor

\State $\varepsilon \gets Threshold \left(Sort\left(\Omega_{IJ}\right)\right)$;

\State 
\State \# \textit{Zeroth Order Constraint}
\For{variable $I,J$}
    \If{$\Omega_{IJ}>\varepsilon$}
        append $\left(I\to J\right)$ to DAG;
    \EndIf
\EndFor

\State 
\State \# \textit{First Order Constraint}
\For{node $I\in$ DAG with in-degree $>1$}
    \State survives $\gets$ True;
    \For{nodes $J,K\in parents(I)$}
        \For{states $i\in I,j\in J,k\in K$}
        \State $\omega_{ijk} \gets |W\left(i\mid j,k\right)|$;
        \If{$\omega_{ijk}>\varepsilon$}
        survives $\gets$ False;
        break;
        \EndIf
        \EndFor
    \If{not survives}
    remove $(I\to J)$ from DAG;
    \EndIf
    \EndFor
\EndFor
\end{algorithmic}
\end{algorithm}

Our algorithm is summarised in the pseudocode Algorithm \ref{alg:cap} and is publicly available through a Python implementation on GitHub.\footnote{In https://github.com/F-Barroso/Inference}

In the case that the threshold is determined as the value above which the network ceases to be connected, the function Threshold can simply take the form of a binary search.
When the knee method is applied, the Threshold function simply represents the application of the method to find the point of greatest inflection.

\subsection{Computational complexity} 
The determination of the threshold requires CI computations between all pairs nodes, which grows as $\mathcal{O}(n^2)$, with $n$ being the number of nodes in the network. Note even though discrete measures, such as NI, require computing CI between states of different nodes, as the number of states per node is finite, those cycles do not grow with network and can thus be disregarded for complexity computations. These are the computations that dominate the determination of the threshold and the zeroth order constraint.

The first order constraint evaluates triads in the network. This phase grows as $\mathcal{O}(t)$, with $t$ representing the number of triads in the network after the first step. If no threshold was established, this is bounded by $\mathcal{O}(n^3)$. This corresponds to a highly dense network, in which the mean degree is of the order of the number of nodes.
However, for the use cases considered, the number of parents for each node is not dependent on the size of the network. In fact, as long as the first and second moments of the degree distribution of the original network are bounded, the number of triads to test in the second step will not grow with the network size. 

This means that, in such cases, the complexity of the first order constraint is $\mathcal{O}(n)$. Thus, the zeroth order dominates the computation time of the algorithm.
The exception would be the case of an outlier weak link in the original network, which would force a low threshold and result in the inclusion of a large number of false positives in the first step. This results in $\mathcal{O}(n^2)$ operations being required in the second step. In such cases, the knee method returns the requirement in the second step to $\mathcal{O}(n)$.

The performance of the second step is thus dependent on the adequate choice of a threshold. Methods to reduce the number of false positives, such as the application of the knee method, greatly increase the efficiency of the second step.

\section{Results for real-world networks}\label{Real networks}
We carried out network inference using our algorithm with the NI measure, for both the Connected and Knee thresholding methods (for brevity we will refer to these as NIConnected and NIKnee). We also repeated the tests using the Fisher z-transformation of the Pearson correlation coefficient (see \cite{kalisch2005}) as the CI measure. This measure is appropriate for continuous data, and is the same measure used in the PC Algorithm we used as a benchmark for our results \cite{zhang2021gcastle} (FisherConnected and FisherKnee, respectively). We present the results of inferring both the skeleton as well as the causal network with directions.

These networks highlight challenges when applying the algorithms of causal inference, explaining some of those outlined in the synthetic data presented in Section \ref{Synthetic networks}.

The quality of results can be characterized in terms of the false positive rate (FPR) and false negative rate (FNR). The FPR is the number of excess edges added by the algorithm which are not present in the original network, normalized by the total number of non-edges in the original network. The FNR is the number of missing edges, that is, edges in the original network which were not identified by the algorithm, normalized by the number of edges in the original network. A measure which combines false positives and false negatives (along with true positives and negatives) is the Matthew's correlation coefficient (MCC) \cite{matthews1975comparison}.

\subsection{ASIA network}
\begin{figure}[!h]
    \centering
    \includegraphics[width=0.8\columnwidth]{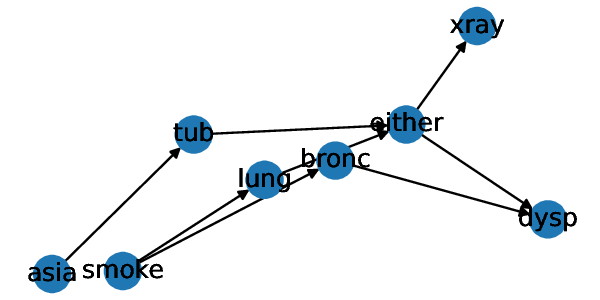}
    \caption{ASIA network. Despite naming the network, the connection from node Asia, Asia$\rightarrow$Turberculosis, is not easy to predict.}
    \label{fig:asia_true}
\end{figure}

ASIA is a small network, commonly used to test Bayesian learning algorithms, proposed by S. L. Lauritzend \cite{Lauritzen1988}, represented in Figure \ref{fig:asia_true}.


Our methods were inferior to the benchmark PC algorithm when inferring both the skeleton and the DAG for the ASIA network, though still presenting reasonable results in most cases, including being faster in all but one case. The results are summarized in Table \ref{table:asiatable}.

The NI measures performed better than Fisher when inferring both the skeleton and DAG. However, they failed to predict the edge Asia to Tuberculosis, a very weak edge that other methods often also fail to predict \cite{thost2021directed}. Incidentally, node Asia is left disconnected from the network when the sole spurious correlation that linked in zeroth order is identified and removed. The algorithm is working as intended, as it correctly removed a spurious correlation. More complex versions of the algorithm that include causal recovery could potentially reinclude dropped edges, but this comes with computational cost.

The NI measures also showed some success in predicting the DAG, though still falling short of the PC algorithm which solely mistook the direction of Smoke to Lung Cancer and added a non-causal bidirectional edge in Asia-Tuberculosis. 

The Fisher measures performed satisfactorily when finding the skeleton, but revealed  a fundamental flaw when inferring the DAG: the metric is too sharp, assigning values corresponding to maximum correlation to all except the weakest correlations, which, for small networks with a large number of strong correlations, prevents the choice of an adequate threshold. This situation highlights the fact that causal measures with greater discrimination are better suited for our algorithms.

We present the values of normalized computation time, but we note that, given the small size of the network, these correspond to few seconds in all cases (in our machine, PC takes around $1.5s$ for Skeleton and $4.5s$ for DAG).

\begin{table*}[h!]\centering
\begin{tabular}{|c|c|c|c|c|}
\hline 
\textbf{Skeleton} & \textbf{Time/PC} & \textbf{FPR} & \textbf{FNR} & \textbf{MCC}\tabularnewline
\hline 
PC & $1.000$ & $0.050$ & $\mathbf{0.000}$ & $\mathbf{0.919}$\tabularnewline
\hline 
NIConnected & $0.516$ & $0.100$ & $0.125$ & $0.750$\tabularnewline
\hline 
NIKnee & $\mathbf{0.308}$ & $\mathbf{0.000}$ & $0.375$ & $0.737$\tabularnewline
\hline 
FisherConnected & $0.741$ & $0.100$ & $0.250$ & $0.650$\tabularnewline
\hline 
FisherKnee & $0.741$ & $0.083$ & $0.250$ & $0.667$\tabularnewline
\hline 
\hline 
\textbf{DAG} & \textbf{Time/PC} & \textbf{FPR} & \textbf{FNR} & \textbf{MCC}\tabularnewline
\hline 
PC & $1.000$ & $\mathbf{0.054}$ & $\mathbf{0.125}$ & $\mathbf{0.748}$\tabularnewline
\hline 
NIConnected & $2.071$ & $0.161$ & $\mathbf{0.125}$ & $0.546$\tabularnewline
\hline 
NIKnee & $0.841$ & $0.071$ & $0.500$ & $0.429$\tabularnewline
\hline 
FisherConnected & $\mathbf{0.179}$ & $0.107$ & $0.750$ & $0.143$\tabularnewline
\hline 
FisherKnee & $0.228$ & $0.107$ & $0.750$ & $0.143$\tabularnewline
\hline 
\end{tabular}
\caption{Values of computation time, normalized by the time for PC algorithm, and values of the FPR, FNR, and MCC for the inferred DAG and skeleton of the ASIA network. In bold we highlight the best results.}\label{table:asiatable}
\end{table*}

\subsection{ALARM network}
Another widely used network for Bayesian learning, representing a medical diagnostic, is the ALARM network \cite{Beinlich1989}.

We obtained very good results for inference using NIKnee, inferring more accurately the skeleton than PC algorithm and inferring the DAG with equivalent results, as measured by the MCC (see \ref{table:alarmtable}), in both cases with significant speedups over the benchmark.

The results for NIConnected, on the other hand, illustrate a shortcoming of the topological threshold that requires connectivity. With this method, two of the variables yield pairwise CIs so low that many false positives are included in the network in zeroth order. These are hard to remove evaluating only CI to a single node, in first order, decreasing the quality of the results and greatly increasing computational time.

\begin{figure}[!h]
    \centering
    \includegraphics[width=\columnwidth]{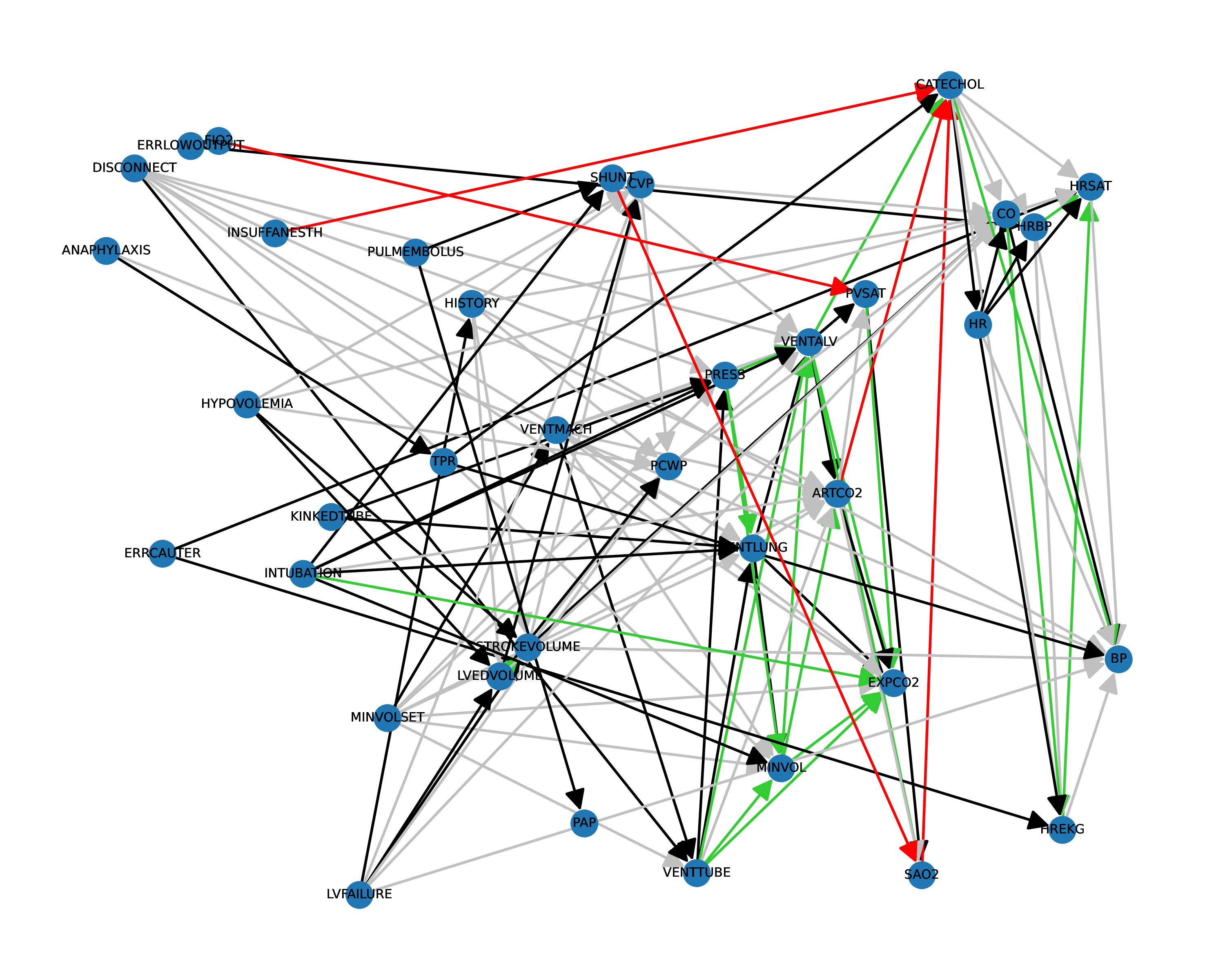}
    \caption{Inferred skeleton of ALARM network using NIKnee. The labels were not displayed in order to avoid cluttering the image. Black lines represent edges correctly predicted, lines in green edges in excess (false positives) and lines in red missing edges (false negatives). Note that two nodes were left disconnected.}
    \label{fig:alarm_knee}
\end{figure}

NIKnee solves the problem by identifying a threshold that removes those variables from the largest connected component. See Figure \ref{fig:alarmgcc}, illustrating the functioning of the Knee method for this network. The effect of including the two weakly connected nodes is visible in the plot of nodes included against edges included in zeroth constraining order. The inferred network from the NIKnee method presents much better MCC results - see Table \ref{table:alarmtable} - yielding the network in Figure \ref{fig:alarm_knee}.

Once again, the Fisher metrics are too aggressive to yield a topological threshold when inferring DAGs. This problem is exacerbated by the fact that these measures are symmetric, therefore when there is an edge that passes the threshold also, the opposite direction will also be included, inflating the FPR. These problem do not occur or are not as evident when inferring the skeleton, thus the measures still present good results in very short computational times. 

\begin{table*}[h!]\centering
\begin{tabular}{|c|c|c|c|c|}
\hline 
\textbf{Skeleton} & \textbf{Time/PC} & \textbf{FPR} & \textbf{FNR} & \textbf{MCC}\tabularnewline
\hline 
PC & $1.000$ & $0.081$ & $0.065$ & $0.625$\tabularnewline
\hline 
NIConnected & $1.100$ & $0.410$ & $\mathbf{0.022}$ & $0.290$\tabularnewline
\hline 
NIKnee & $0.389$ & $\mathbf{0.016}$ & $0.174$ & $\mathbf{0.794}$\tabularnewline
\hline 
FisherConnected & $0.248$ & $0.134$ & $\mathbf{0.022}$ & $0.543$\tabularnewline
\hline 
FisherKnee & $\mathbf{0.106}$ & $0.066$ & $0.304$ & $0.508$\tabularnewline
\hline 
\hline 
\textbf{DAG} & \textbf{Time/PC} & \textbf{FPR} & \textbf{FNR} & \textbf{MCC}\tabularnewline
\hline 
PC & $1.000$ & $0.043$ & $0.174$ & $\mathbf{0.555}$\tabularnewline
\hline 
NIConnected & $0.551$ & $0.338$ & $\mathbf{0.022}$ & $0.241$\tabularnewline
\hline 
NIKnee & $0.244$ & $\mathbf{0.036}$ & $0.239$ & $0.547$\tabularnewline
\hline
FisherConnected & $0.177$ & $0.161$ & $\mathbf{0.022}$ & $0.376$\tabularnewline
\hline 
FisherKnee & $\mathbf{0.035}$ & $0.064$ & $0.674$ & $0.181$\tabularnewline
\hline 
\end{tabular}
\caption{Values of time, normalized by the time for PC algorithm, FPR, FNR, and MCC for the inferred ALARM network. In our machine, PC registered times of $462s$ for Skeleton and $1136s$ for DAG. In bold we highlight the best results.}\label{table:alarmtable}
\end{table*}


\section{Results for synthetic networks}\label{Synthetic networks}

In this section, we present a more general study of network inference, analysing larger and denser networks by testing our algorithm for synthetic networks of varying size, and compared the results with those obtained using the PC algorithm, both for skeleton and DAG inference.

We evaluated the performance for two network topologies with discrete data. The first method (1.) generated networks adding nodes recursively from a chosen sink, selecting in- and out-degrees from a narrow statistical distribution, similar to a Poisson. This minimizes correlations between temporal order of the node and node degree, as well as between node in- and out-degrees. By construction, networks generated by this method have a single sink, thus representing the causal past of a single effect. However, the causal identification method works just as well with multiple sinks.

The second method (2.) generated networks from random triangular adjacency matrices. This approach was used for example in \cite{kalisch2005}. Although total node degrees are distributed narrowly around the mean value, this approach does result in anticorrelations between node in- and out-degrees, as well as correlations of both with node order.

For these two methods, 1. and 2., once the network is generated, between 2-4 states are randomly attributed to each node. Then, we generate random conditioned probabilities according to the Markov decomposition of the DAG, and finally generate data according to these probabilities. These methods yield discrete data from which we seek to infer the original network.

\begin{figure}[h!]
    \centering
    \includegraphics[width=\columnwidth]{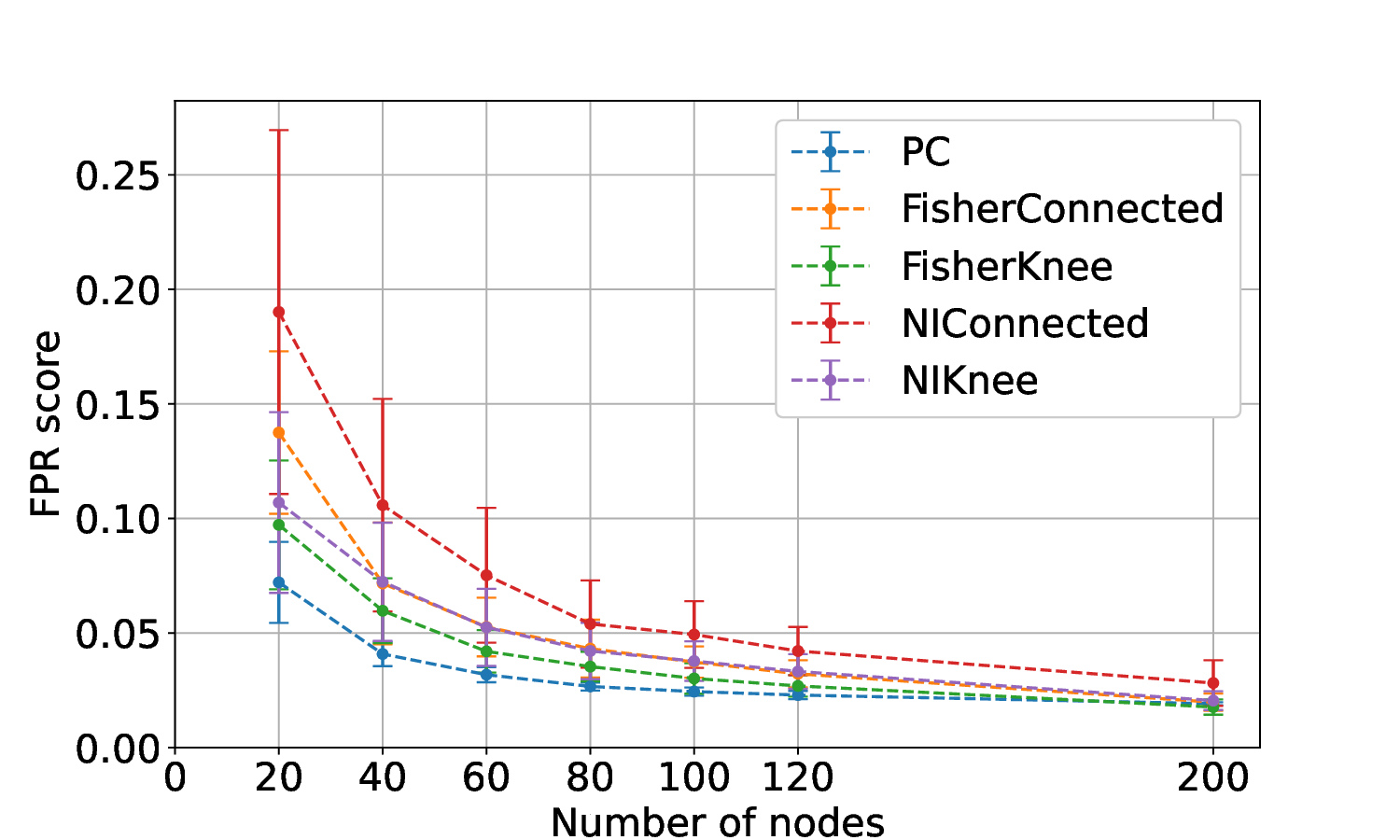}
    \caption{Variation of FPR score for inferred DAG with network size, for networks generated with mean degree close to $3$ and by method 2. Bars represent the range of results (and not uncertainty in the mean value). Each point corresponds to a mean of $30$ repetitions, except for nodes $120$ and $200$ which were computed with $22$ and $9$ repetitions, respectively.}
    \label{fig:fpr}
\end{figure}
\begin{figure}[h!]
    \centering
    \includegraphics[width=\columnwidth]{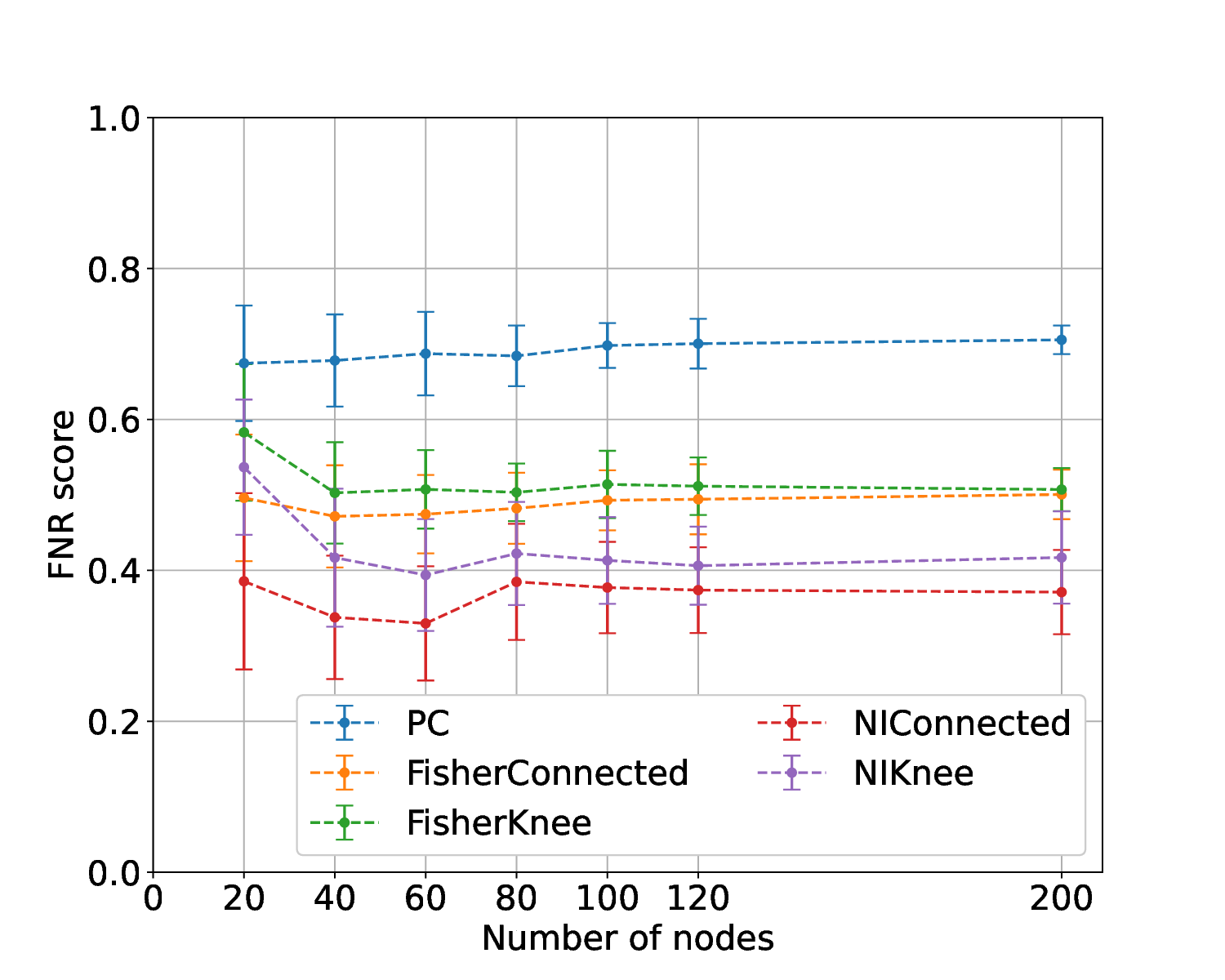}
    \includegraphics[width=\columnwidth]{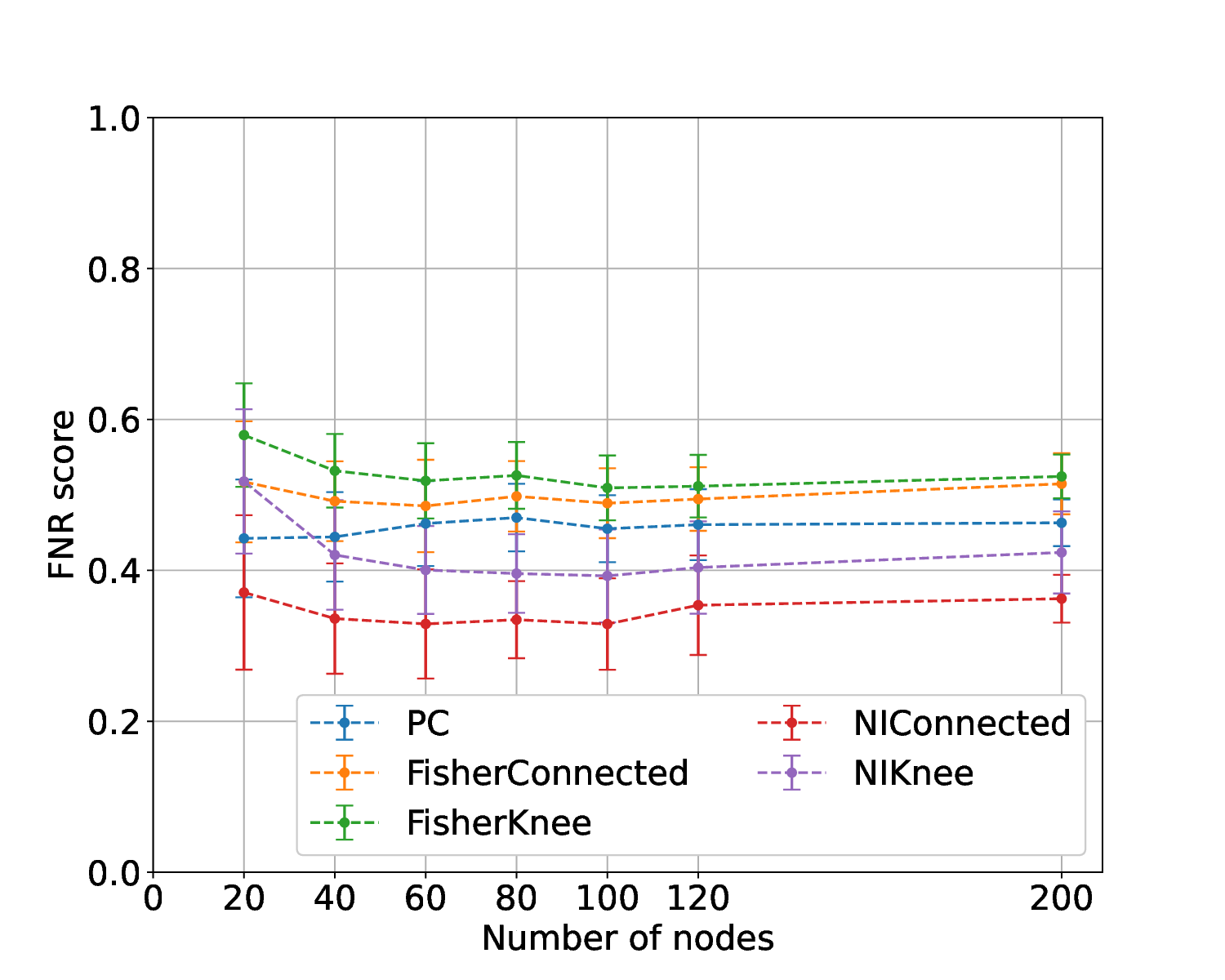}
    \caption{Top: Variation of FNR score for inferred DAG with network size, for mean degree close to $3$ and networks generated by method 2. The number repetitions is as above. Bottom: Variation of FNR score for inferred Skeleton in the same conditions. Each point corresponds to, respectively, in increasing order, means of $54$, $41$, $40$, $39$, $24$, $20$, and $5$ repetitions.}
    \label{fig:fnr}
\end{figure}
\begin{figure}[h!]
    \centering
    \includegraphics[width=\columnwidth]{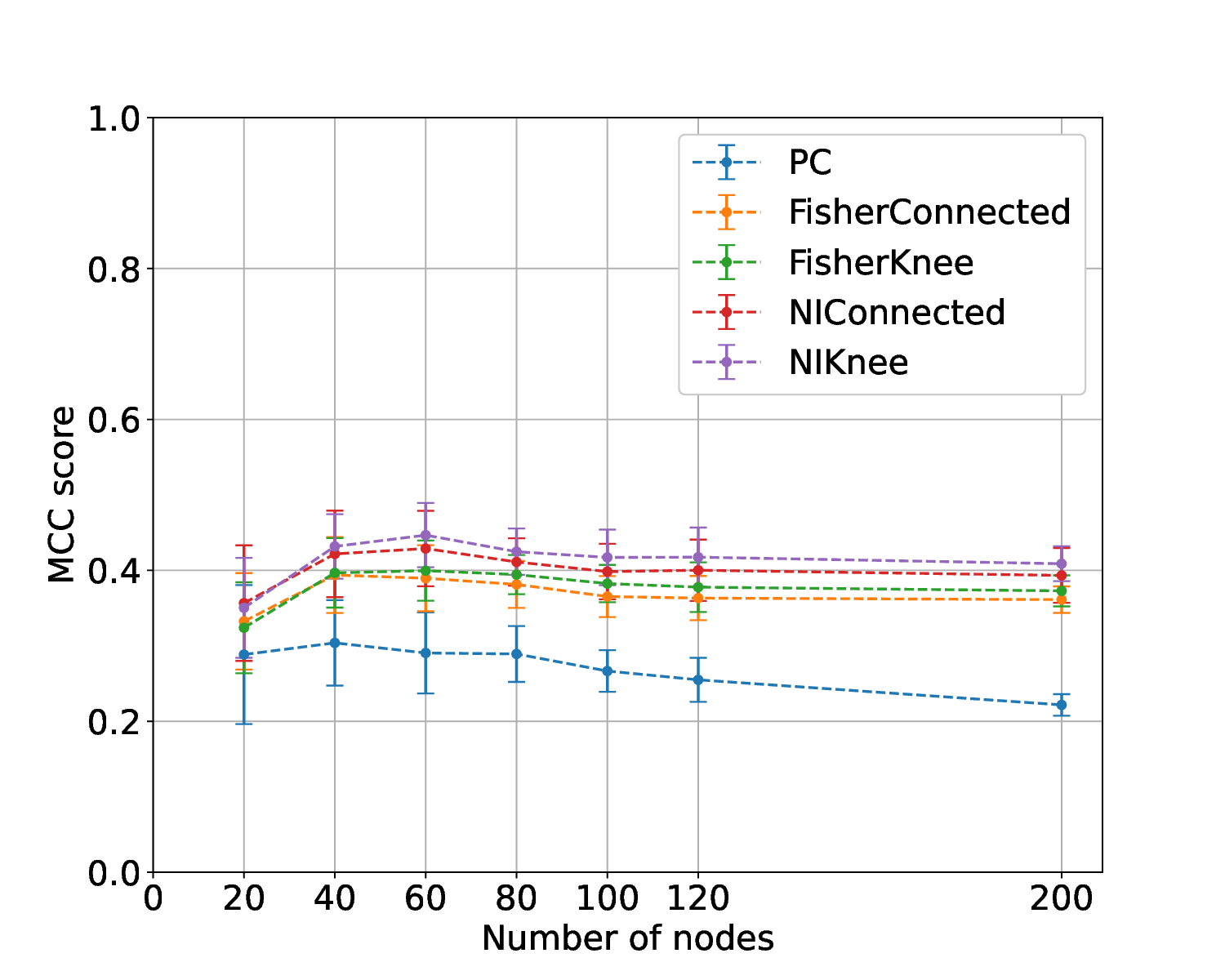}
    \caption{Variation of MCC score for inferred DAG with network size, for mean degree close to $3$, for networks generated by method 2. The number of repetitions is the same as in Figure \ref{fig:fpr}.}
    \label{fig:mcc}
\end{figure}
\begin{figure}[h!]
    \centering
    \includegraphics[width=\columnwidth]{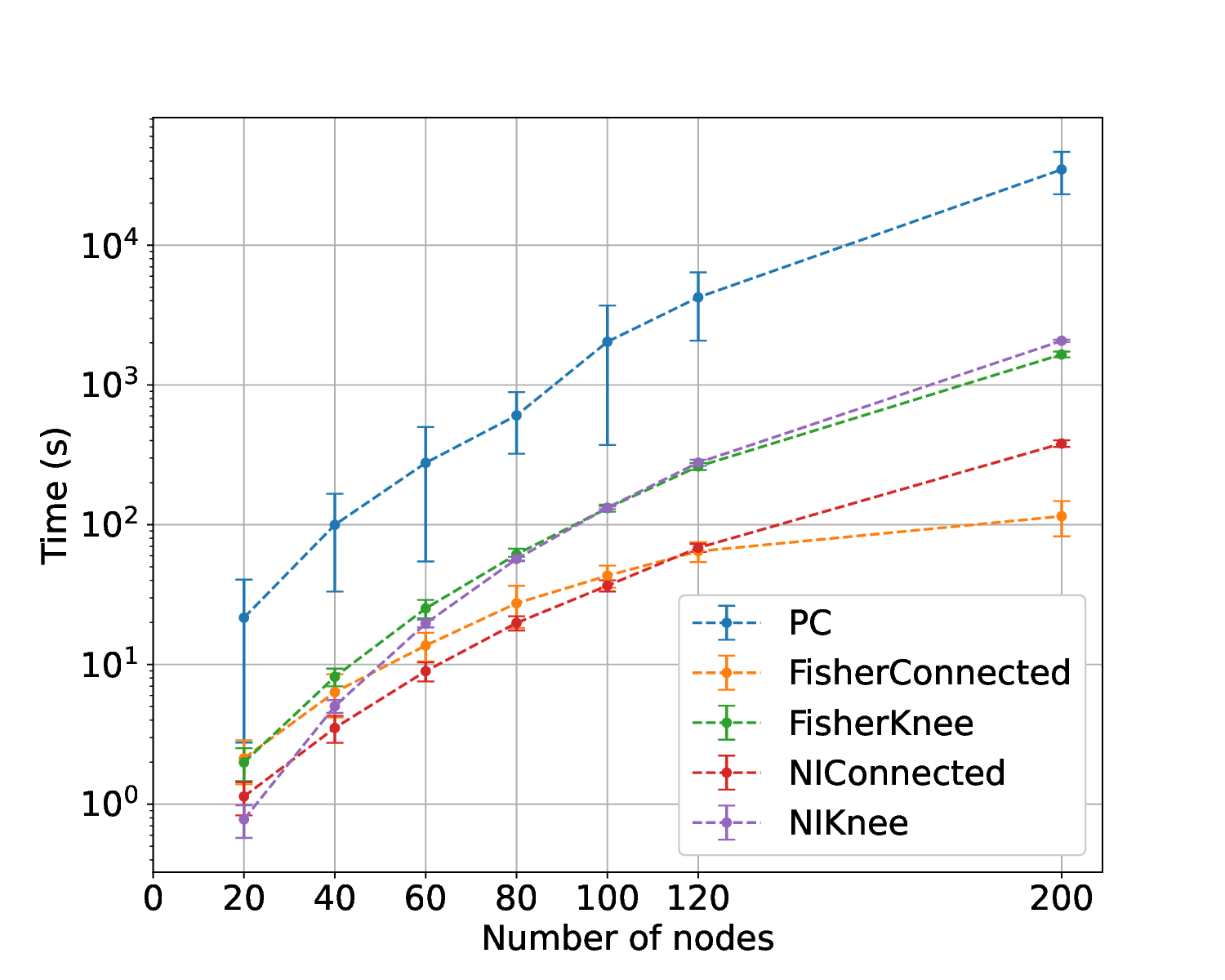}
    \caption{Variation of computation time for inferred DAG with network size, for networks generated by method 2. The number of repetitions is the same as in Figure \ref{fig:fpr}.}
    \label{fig:time}
\end{figure}

Results for FPR, FNR, and MCC for a range of network sizes with mean degree close to $3$, generated by method 2., are shown in Figures~\ref{fig:fpr}, \ref{fig:fnr}, and \ref{fig:mcc}. Inference was performed from $10^4$ lines of data. Computational time are compared with those for the PC algorithm in \ref{fig:time}. We observe that for discrete data, both algorithms, Connected and Knee, provide very good results.

Both PC and our algorithm, independent of measure or threshold used, present FPRs that decrease with the number of nodes. This behaviour is witnessed both when inferring DAGs or the skeletons, for either method, by virtue of the large number of true negatives which dominate the normalization factor for large networks. It shows that both algorithms are generally very good at excluding connections between clearly independent variables. NIConnected performs slightly worse than the other methods regarding FPR for networks generated by method 2. (see an example in Figure \ref{fig:fpr}), while the PC algorithm performs slightly worse then the remaining methods when inferring skeletons.

The main differences are registered for FNR (see Figure \ref{fig:fnr}). NIConnected and then NIKnee perform generally better, both for skeleton and DAG inference, with the trend being clearer for larger networks. Fisher measures have similar performances, with around $0.1$ higher FNRs than NI measures. The PC algorithm presents FNR generally slightly worse than NIKnee, when inferring skeletons of networks generated by method 1. and 2., respectively.
We performed the same tests for less dense, thus more tree-like, networks - we chose a mean degree close to $1.3$. In this case, the gap is even clearer. PC performs significantly worse than all the other measures when inferring DAGs, for either method (over $0.1$ higher than the Fisher measures).

These two metrics combine to an MCC for all four variations of our algorithm that are significantly better than PC's for larger networks. NIKnee performs generally the best closely followed by NIConnected in networks generated by 1. or the Fisher metrics in networks generated by 2. However, a notable exception are the MCC scores of NIConnected, which are as low as PC for networks generated by 2. with low degree. We show in Figure \ref{fig:mcc} the results of MCC for inferred DAG with the remaining methods presenting similar behavior, except when otherwise stated.

In terms of time, our algorithm outperformed PC algorithm by approximately an order of magnitude, whether using Fisher or NI measure (see Figure \ref{fig:time}). The Fisher measures are generally faster than NI measures for larger networks, as they do not require cycles over the combination of states. The Connected metrics are generally faster than the Knee metrics, by virtue of using binary search, which cannot be applied to the latter, to find the location of the threshold.

When analysing the number of removed spurious correlations in each step when inferring DAGs of size $60$, mean degree $3$, generated by method 2., we found out in the zeroth step, a mean of $92\%$ of the spurious edges are removed (or alternatively, return a network with $8\%$ of spurious edges) for the Connected methods, while the Knee methods remove a mean of $94\%$ of spurious edges (that is, return a network with $6\%$ spurious edges). On top of this, $97-98\%$ of the remaining spurious correlations are removed in the first step by the Fisher measures, while the NI measures remove $99\%$ of the remaining spurious edges. These values correspond to means of $101$ networks. Furthermore, we note that the zeroth step contributes to $95\%$ of the FPR. Regarding wrongly removed edges in the first step, in around half the tries, the Fisher metrics wrongly removed $1$ or $2$ directed edges, in some cases removing as much as $8-9$ edges. NI measures performed much better in this regard, wrongly removing edges in only $6\%$ of the runs, never more than a single one. Therefore, we consider the sequential approach validated.



\begin{figure}[h!]
    \centering
    \includegraphics[width=\columnwidth]{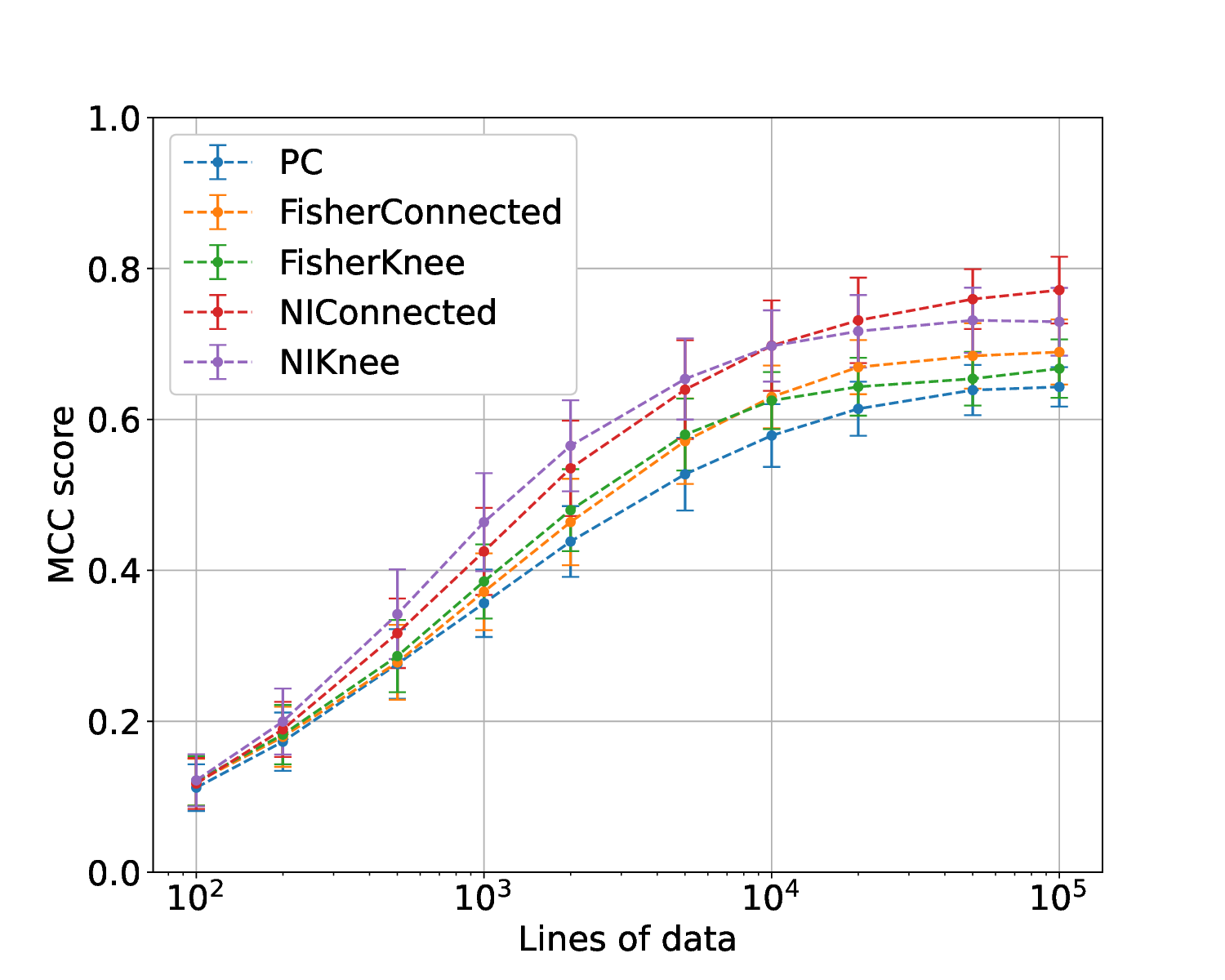}
    \caption{Variation of the MCC score for inferred Skeleton with the amount of data for networks generated by method 1. Each point corresponds to a mean of $56$ repetitions.}
    \label{fig:depleted_mcc}
\end{figure}

A question of special importance is the amount of data needed for functionality of the algorithm. We analysed and compared the behaviour of the algorithms for networks of $60$ nodes, mean degree close to $3$ with varying amount of data. The MCC scores for all measures, both when inferring the DAG or skeleton, decrease with smaller data samples. We observe that NI measures still perform generally better than PC algorithm, even for small amounts of data, with the difference becoming more pronounced with more data (see Figure \ref{fig:depleted_mcc} for an example inferring the DAG). We start observing convergence in the quality results for sample sizes of around $10^4$, which is the sample size used in the previous subsection, for method 1., to $10^5$, for method 2.

\begin{figure}[h!]
    \centering
    \includegraphics[width=\columnwidth]{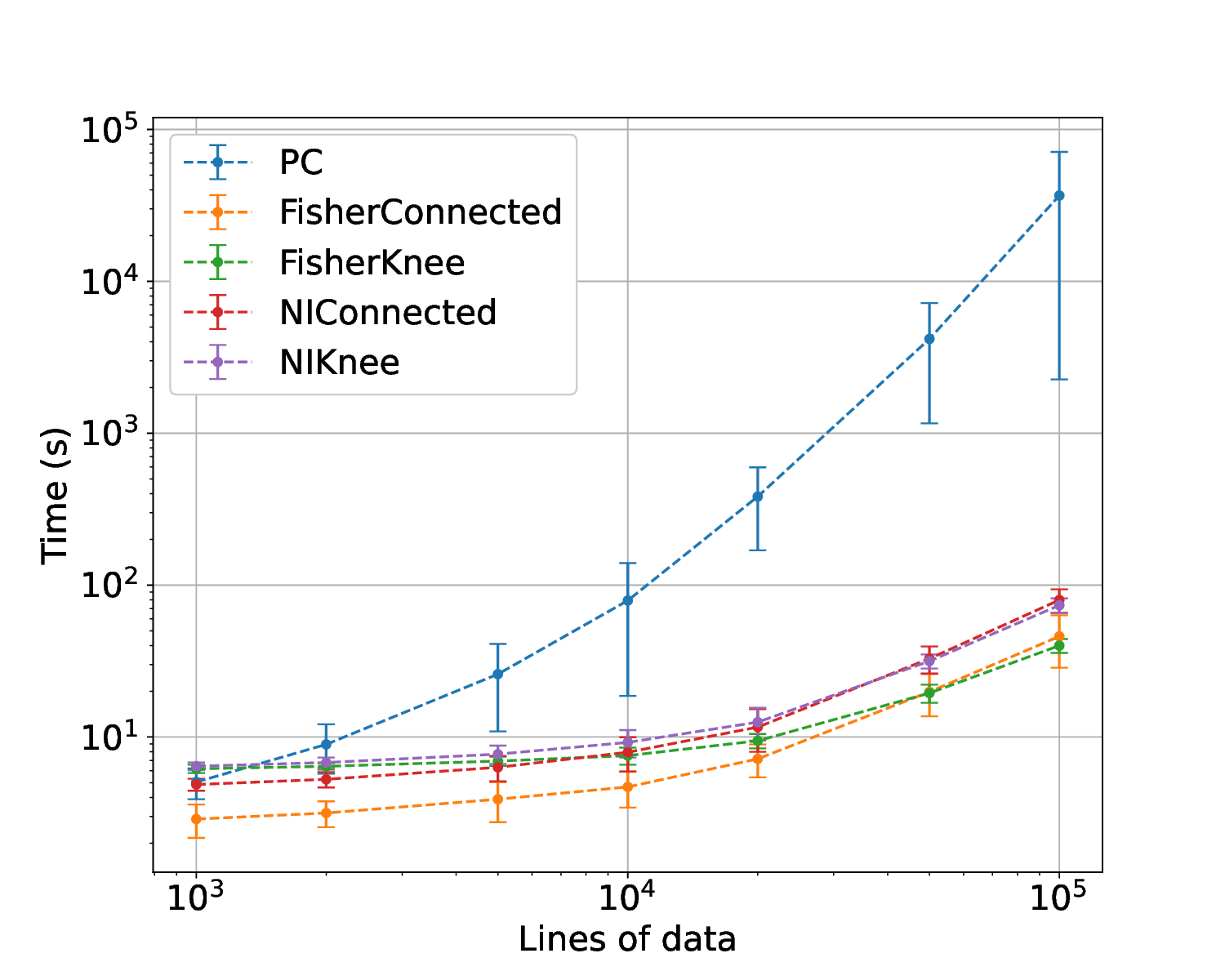}
    \caption{Variation of computation time, in seconds, for inferred Skeleton with the sample size, for networks generated by method 1. Each point corresponds to a mean of $56$ repetitions.}
    \label{fig:depleted_time}
\end{figure}

Although the PC (and FisherConnected) algorithm are generally very fast for data samples until size $10^3$, we notice that the algorithm scale very badly with sample size, becoming as high as two orders of magnitude slower than our proposed algorithms for $10^5$ lines of data (see Figure \ref{fig:depleted_time} for an example inferring the DAG), when convergence starts to be observed.

\begin{figure}[h!]
    \centering
    \includegraphics[width=\columnwidth]{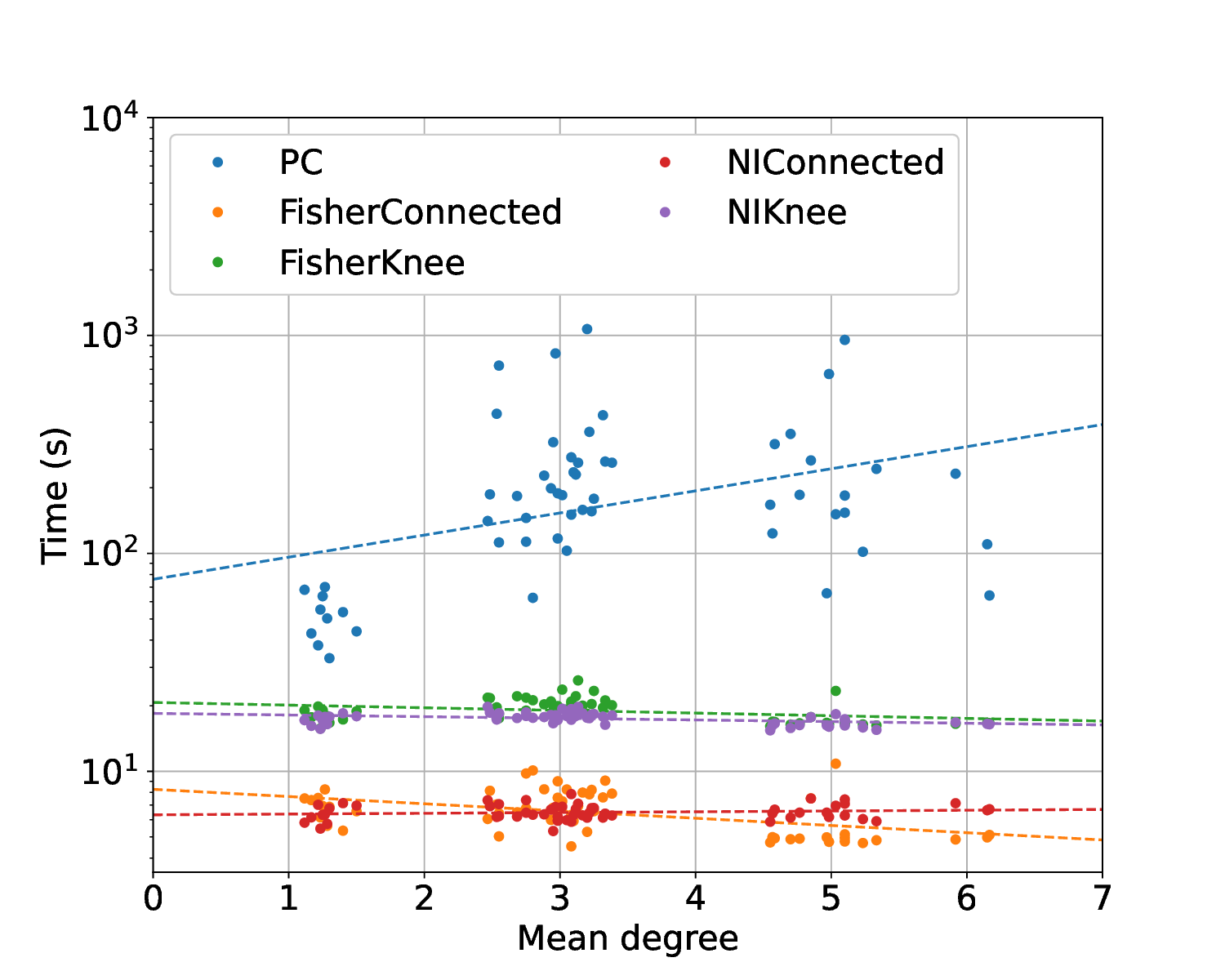}
    \caption{Variation of time, in seconds, for inferred DAG with the mean degree, for networks generated by method 1., with a total of $56$ data samples. The dashed lines correspond to linear regressions, though it is not a good model for PC's behaviour.}
    \label{fig:meandeg_time}
\end{figure}

\begin{figure}[h!]
    \centering
    \includegraphics[width=\columnwidth]{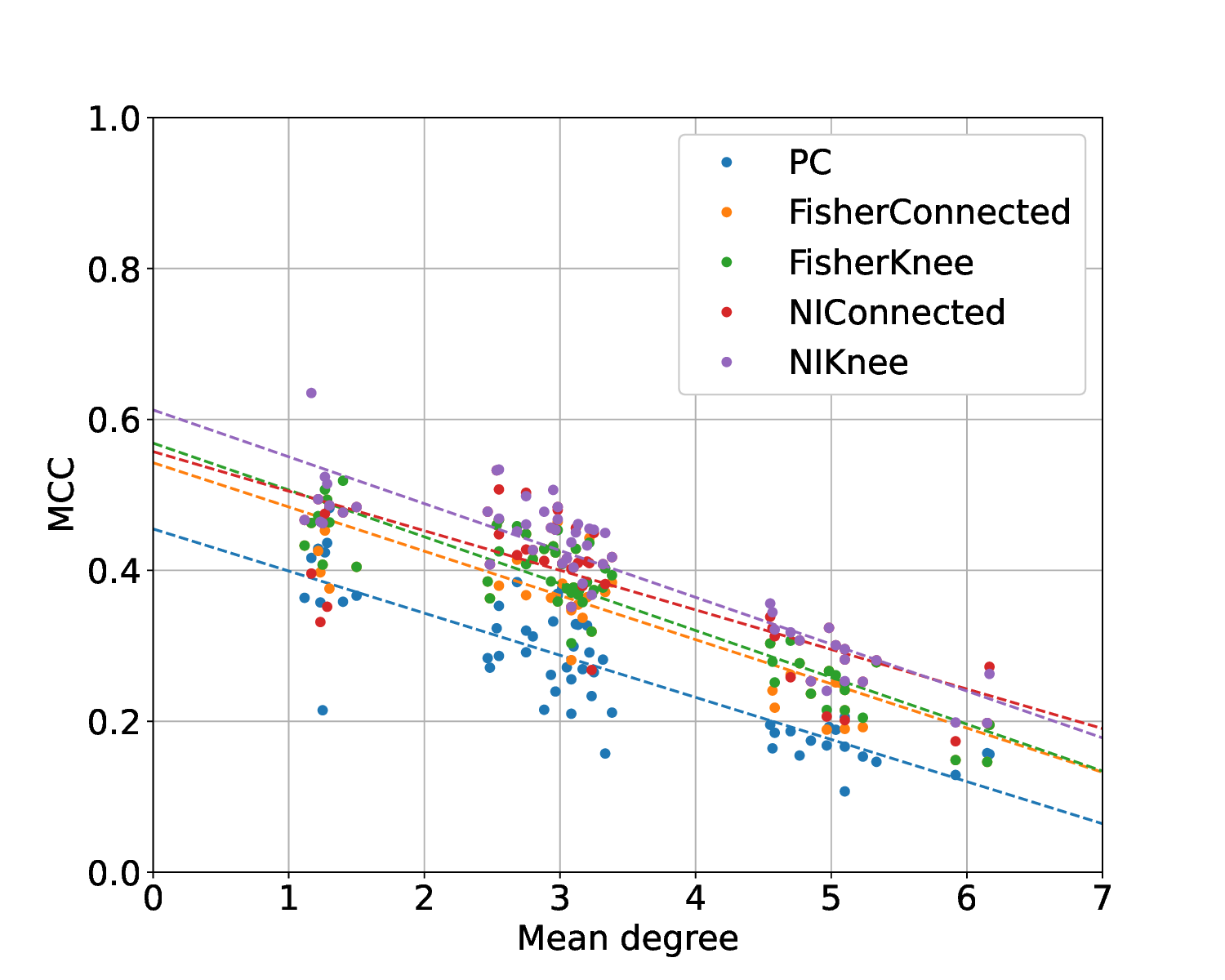}
    \caption{Variation of MCC for inferred DAG with the mean degree, for networks generated by method 1., with a total of $56$ data samples. The dashed lines correspond to linear regressions.}
    \label{fig:meandeg_mcc}
\end{figure}

Finally, we present in Figures \ref{fig:meandeg_time}, \ref{fig:meandeg_mcc} the behaviour of all metrics with time and MCC, respectively, with variations mean degree for networks of size $60$ generated by method 2., for inferred DAG. Results for method 1., and inferred skeleton are similar. The time is approximately does not vary with increase in mean degree, highlighting the fact that the zeroth step dominates the complexity of the algorithm.

The variation of MCC with mean degree is approximately the same for all measures, indicating that there our proposed algorithm are superior to the benchmark for any density of the theoretical network. This is of particular importance, as the density of the causal network behind the data is hard to estimate \emph{a priori}, complicating the choice of an adequate algorithm if it were only superior in a set densities.

\section{Discussion and Conclusions}\label{Conclusion}
In this article, we have proposed a novel constraint-based algorithm for inferring causal networks, which establishes automatic topological thresholds computed from the data. We have proposed two alternatives: one that seeks to connect all nodes in a single component, and another which seeks the optimal compromise between number of nodes in the largest connected component and edges added to the network with values above the threshold. The later is computed finding the knee of the described curve.

We also introduced a novel conditional dependence measure, Net Influence, similar to the certainty factor, which allows for the calculation of state-by-state influence. We have shown it generally performs very well with our algorithm, both when inferring the skeleton and the directed causal network (DAG).

Our algorithm (especially using the NI metric) can be used to directly infer directionality of the edges, whereas in algorithms such as PC it is done as a second step after inferring the skeleton. This speeds up the process. Furthermore, the asymmetry in NI captures information that is not utilized by other algorithms.

We compared our algorithm for discrete data with a benchmark, PC algorithm, evaluating the quality of results and running time. We tested our algorithm for both proposed topological thresholds, using our NI metric and, for comparison, the Fisher correlation.
When testing on two real networks we obtained mixed results. PC algorithm generally outperformed our methods in a very small network, ASIA. For a larger network, ALARM, our algorithm outperformed PC for the knee threshold, but failed to produce acceptable results for the connected threshold. This is due to the presence of a very weakly connected node, which forces the threshold to be very low, resulting in a large FPR. The knee algorithm avoids this problem.


These results were then generalized for even larger networks, randomly generated and fitted with synthetic data. Our method outperformed, both in quality and time, the PC algorithm, with the difference being clearer for larger networks. This trend is generally independent of the mean degree of the original network so our method is capable of presenting reliable and very fast results compared to the benchmark. Convergence of the quality results starts to be observed for around $10^4$ lines of data, though the trend of our algorithm outperforming PC is observed even for lower data magnitudes.

One might expect that the topological threshold method we have described, based on the connectivity of the inferred network, could be biased towards less dense networks. We did find a decrease in the quality of results with increasing network density, however this decline was also observed in the reference PC algorithm, such that our algorithm maintained its advantage.

We observed similar results for the four tested variations of our algorithm, using two different metrics combined with two different threshold determination algorithms. The results for our NI measure were slightly superior to those using the Fisher measure. In general, the Knee method obtained lower false positive rates, while the more conservative Connected method resulted in lower false negative rates. The Knee method does have a higher computational cost, but on the other hand, the Connected method does not adapt to cases in which very weak connections are present, resulting in poorer performances. Unless one is sure that all variables must be connected to a single causal network component, then, we suggest that the more consistent and generally best performing NIKnee method should be used by default.

We have shown that the conditional independence testing can be carried out in increasing order of complexity, testing first unconditioned interactions (zeroth order) then conditioned on one other variable (first order) and so on. 
We found that only up to first order conditioning was sufficient to produce excellent results, while keeping the computation time low. This is the most important tradeoff we found in increasing the algorithm speed.
These speed advantages only increase with system size. Our algorithm was generally at least ten times faster than the standard PC algorithm across a range of cases.

Our simple method can be easily implemented and applied to very large datasets. The use of state-wise comparisons, rather than correlations or mutual information across all the states of a variable, reflects the structure of many Bayesian networks, and facilitates the use of the inferred network to analyse the evolution of the system, or to infer which state may be the root cause of a given outcome.

\bibliography{references}{}
\bibliographystyle{unsrt}

\end{document}